\pdfoutput=1
\def\CLASSINPUTinnersidemargin{0.75in}
\def\CLASSINPUToutersidemargin{0.75in}
\def\CLASSINPUTbottomtextmargin{0.75in}
\documentclass[conference]{IEEEtran}
\IEEEoverridecommandlockouts
\usepackage{cite}
\usepackage{amsmath,amssymb,amsfonts}
\usepackage{graphicx}
\usepackage{url}
\usepackage{xcolor}
\usepackage{booktabs}

\title{\vspace{0.25in}Accelerating Optimization over Graphs of Convex Sets via Neural Network Approximations}
\author{Ananya Trivedi$^{1}$, Sarvesh Prajapati$^{1}$, Zhexin Xu$^{1}$, Mohamed Khalid M Jaffar$^{2}$,  \\ David Rosen$^{1}$, and Ta\c{s}k{\i}n Pad{\i}r$^{1,3}$ %
\thanks{$^{1}$Northeastern University, Boston, Massachusetts, USA.}
\thanks{$^{2}$University of Maryland, College Park, Maryland, USA.}
\thanks{$^{3}$Ta\c{s}k{\i}n Pad{\i}r holds concurrent appointments as a Professor of Electrical and Computer Engineering at Northeastern University and as an Amazon Scholar. This paper describes work performed at Northeastern University and is not associated with Amazon.}
\thanks{Corresponding author: \tt\small trivedi.ana@northeastern.edu}
}

\begin{document}
\maketitle

\begin{abstract}
Motion planning problems such as collision-free navigation and contact-rich manipulation can be naturally formulated as optimization problems that couple discrete decisions with continuous trajectories. The Graphs of Convex Sets (GCS) framework offers a practical solution to these problems. It represents discrete decisions as nodes of a graph and  encodes continuous trajectories in the edges connecting them. However, the resulting optimization subproblems can become computationally prohibitive for online replanning. 

In this work, we propose a learning-based strategy to mitigate this limitation. Specifically, we replace the costly convex relaxation step required by nominal GCS with a single forward pass through a Graph Attention Network that predicts a set of highly probable candidate paths through the graph. A lightweight ranking network then orders these candidates by their estimated trajectory cost. Evaluating them in this order, we terminate our search early while still recovering a near-optimal motion plan. We validate the resulting pipeline across diverse robotic tasks, including collision-free motion planning for a 3D quadrotor and a 7-DoF manipulator, and planning through contact for planar pushing. Across both convex and non-convex cost and constraint settings, our approach yields up to two orders of magnitude speedup over nominal GCS while maintaining a 100\% success rate, at the cost of some suboptimality in the recovered solutions. Code implementations and video demonstrations can be found at \textcolor{magenta}{\url{https://neural-gcs.github.io/}}.
\end{abstract}
\begin{figure*}[t]
\centering
\includegraphics[width=0.9\textwidth]{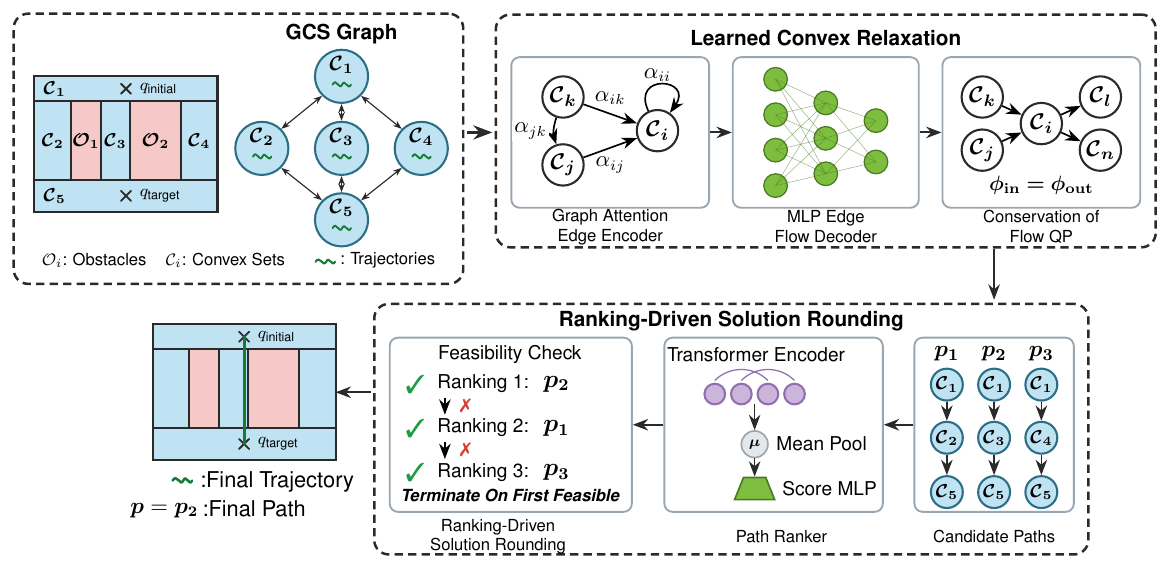}
\caption{Overview of the proposed learning-based strategy to accelerate GCS.}
\label{fig:block_diagram}
\vspace{-4mm}
\end{figure*}
\vspace{-3mm}
\section{Introduction}
\label{sec:introduction}
Many motion planning problems require making discrete and continuous decisions simultaneously. In the case of a quadrotor navigating a cluttered environment, the planner must decide which side of each obstacle to pass while shaping a smooth trajectory along the chosen route. For legged locomotion, the planner must select a sequence of footstep locations and simultaneously optimize for the whole-body motion that keeps the robot balanced. Indeed, seemingly disparate tasks such as collision-free motion planning \cite{gcs_planning_paper_original} and hybrid systems trajectory optimization \cite{gcs_planning_through_contact} share a similar underlying structure: a discrete choice tightly coupled to a continuous trajectory.

Classical approaches to problems of this form involve solving mixed-integer programs \cite{footstep_planning_mixed_integer_russ}. While these techniques recover globally optimal solutions, the computation times do not scale well to long-horizon planning problems involving high-dimensional state spaces \cite{posa_cito}. Graphs of Convex Sets (GCS) \cite{gcs_optimiazation_paper_original} is a framework that addresses this limitation by reformulating the problem as a shortest path problem (SPP) on a directed graph whose vertices are paired with convex sets. This abstraction maps naturally onto motion planning. In collision-free navigation, for instance, the convex sets represent obstacle-free regions of the configuration space. GCS jointly optimizes for the selection of these regions and a dynamically feasible trajectory through them.

In practice, GCS solves this SPP by first computing an empirically tight convex relaxation of the underlying mixed-integer program. The rounding stage then samples multiple candidate paths based on the relaxation output, evaluates each by solving an optimization problem, and returns the best feasible motion plan \cite{gcs_planning_paper_original}. This amounts to solving a few optimization problems whose computational cost is small compared to the branch-and-bound techniques needed for the mixed-integer formulation. Despite this speedup, the solve times for many practical planning problems are not yet suited for online replanning. For instance, planning a quadrotor trajectory  with GCS takes on the order of a few seconds. For contact-rich tasks, where bilinear dynamics introduce inherent non-convexities, solve times can stretch to minutes.

The present work is motivated by the need to reduce the computation time of GCS. To this end, we first generate an offline dataset of GCS solutions by randomizing start and goal configurations and the underlying graph structure. We then train a Graph Attention Network (GAT)~\cite{graph_attention} to approximate the output of the convex relaxation stage. This choice is motivated by the ability of GAT to learn long-range dependencies over graph-structured data. During online planning, a forward pass through this network replaces the need to explicitly solve the convex relaxation. Using the same dataset, we then train a ranking network that scores the candidate paths generated during the rounding stage by their predicted optimality. By evaluating the top-ranked candidates first and terminating at the first feasible solution, we avoid solving an optimization problem for every candidate.

To demonstrate the generality of our approach, we evaluate it across a wide spectrum of problems GCS is designed to handle, ranging from collision-free 3D quadrotor and 7-DoF manipulator motion planning to contact-implicit planar pushing. Across all of these robotic domains, we explicitly tackle both convex and non-convex cost and constraint formulations. Despite the significant differences in dynamics and constraints between these environments, the same learning-based pipeline we propose yields up to two orders of magnitude speedup over nominal GCS while maintaining a 100\% success rate. These results demonstrate that offline learning can accelerate GCS toward online replanning. Fig.~\ref{fig:block_diagram} provides an overview of the proposed method.
\section{Related Work}
\label{sec:related_work} 
While the GCS framework generalizes to a wide variety of combinatorial optimization problems~\cite{gcs_is_general}, our experimental results specifically benchmark its performance on motion planning tasks. Consequently, we restrict the scope of this review to established motion planning paradigms and recent computational enhancements to GCS for motion planning.\vspace{1mm} \\
\textbf{Optimization-Based Methods: }Framing collision-free motion planning as an optimization problem often introduces non-convexities from obstacle-avoidance constraints, making it difficult to solve the problem to global optimality \cite{motion_planning_1}. GCS, on the other hand  produces certifiably safe motion plans entirely through convex optimization \cite{gcs_planning_paper_original}.
Similarly, contact-implicit trajectory optimization is highly sensitive to the initial guess owing to the non-convexity of linear complementarity constraints \cite{aykut_cito}. GCS, on it's part, reasons globally over contact dynamics, removing the need for a good initial guess \cite{gcs_planning_through_contact}. Our learning-based acceleration further reduces GCS solve times from a couple of minutes to around half a second for contact-rich tasks.\vspace{1mm} \\
\textbf{Sampling-Based Methods: }Rapidly-exploring Random Trees and Probabilistic Roadmaps are widely used for collision-free motion planning \cite{lavelle}, and have been recently extended to contact-rich manipulation \cite{pang_contact} and legged locomotion \cite{sleiman_locomanip}. Unlike GCS \cite{gcs_planning_paper_original}, however, these methods converge only asymptotically to the optimal trajectory.\vspace{1mm} \\
\textbf{Learning-Augmented Methods: }Graph Attention Networks (GATs) \cite{graph_attention} capture long-range dependencies between nodes and edges, and have been used to guide sampling-based planners that operate on graph-based data structures \cite{gnn_planning_upenn}. Trained offline, these networks steer online graph construction toward faster goal convergence, but still inherit the limitations of sampling-based planners and require online collision checking at each node expansion. Instead, we use GATs to approximate the convex relaxation stage, which dominates GCS solve time. Furthermore, because the GCS framework relies on obstacle-free convex regions that are precomputed offline\cite{iris_speedup_gpu}, it inherently eliminates the need for online collision checking.

Beyond collision-free planning, recent techniques mitigate the computational cost of contact-implicit trajectory optimization by decomposing the problem into offline learning of a contact schedule and online optimization of the contact wrench. However, this offline learning relies on hand-crafted heuristics for object manipulation \cite{mcts_for_manipulation} and gaited locomotion \cite{footstep_planning_diffusion}, limiting generality across domains. In contrast, our framework requires no hand-crafted heuristics, applying the same neural network architectures across all problems considered in this work.\vspace{1mm} \\
\textbf{GCS-Based Methods: }
Recent methods to accelerate GCS, such as multi query GCS~\cite{gcs_speedup_savva_offline_sdp} and IxG* GCS~\cite{gcs_speedup_ixg}, rely on precomputing cost-to-go lower bounds offline. This tightly couples their performance to explicit algebraic structures, strictly requiring quadratic costs or purely convex set geometries. Consequently, they are restricted to collision-free navigation and, unlike our method, cannot natively handle the non-convex constraints of contact-rich tasks.

Other efforts such as \cite{gcs_speedup_gcs_star} and \cite{astar_gcs} target scalability to larger graphs. Separately, \cite{iris_speedup_gpu} accelerates the obstacle-free convex region decomposition, which can itself be a bottleneck in high-dimensional configuration spaces. Since our approach uses GCS solutions as a backbone to train neural networks, integrating it with these methods should yield compounding speedups.
\begin{figure*}[t]
\centering
\includegraphics[width=0.9\textwidth]{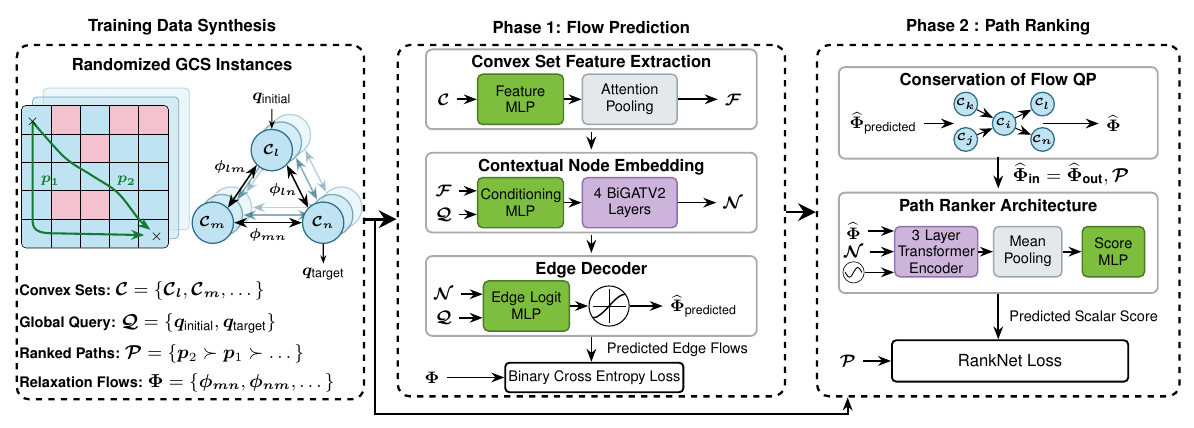}
\caption{Overview of data synthesis and the training pipeline. Solving randomized GCS instances yields convex relaxation edge flows and discrete paths, which serve as ground-truth to supervise the Phase 1 flow predictor and Phase 2 path ranker.}
\label{fig:training_diagram}
\vspace{-4mm}
\end{figure*}
\section{Preliminaries} 
\label{sec:preliminaries} 
Motion planning using GCS~\cite{gcs_planning_paper_original} is equivalent to solving a Shortest Path Problem (SPP) over a directed graph $G=(V, E)$ with vertices $V$ and edges $E$, where each vertex $v \in V$ is associated with a convex set $\mathcal{C}_v$. Solving this problem requires simultaneously determining two interdependent components. The first is the discrete path $p \subseteq V$ connecting a source vertex $s$ to a target vertex $t$. The second is the continuous trajectory $x$ along that path, connecting the initial robot configuration $q_{\text{initial}}$ to the target configuration $q_{\text{target}}$. This trajectory is obtained by optimizing the vertex location $x_v$ within the convex set $\mathcal{C}_v$ of each vertex on path $p$. This can be expressed as the mathematical program Problem~\eqref{eq:spp_in_gcs}:
\vspace{-2mm}
\begin{subequations}
\label{eq:spp_in_gcs}
\begin{align}
\underset{\phi,\,x}{\text{minimize}}\quad 
& \sum_{e=(u,v)\in\mathcal{E}_p} c(x_u,x_v) 
\label{eq:spp_cost}\\
\text{subject to}\quad 
& \mathcal{E}_p = \{e \in E : \phi_e = 1\}, 
\label{eq:spp_active_set}\\
& \phi_e \in \{0,1\}, \quad \forall e \in E, 
\label{eq:spp_binary}\\
& q_{\text{initial}} \in \mathcal{C}_s,\ \ q_{\text{target}} \in \mathcal{C}_t, 
\label{eq:spp_endpoints}\\
& x_v \in \mathcal{C}_v, \quad \forall v \in p, 
\label{eq:spp_containment}\\
& \sum_{(s,w)\in E}\phi_{(s,w)} = \sum_{(u,t)\in E}\phi_{(u,t)} = 1, 
\label{eq:spp_injection}\\
& \sum_{(u,v)\in E}\phi_{(u,v)} = \sum_{(v,w)\in E}\phi_{(v,w)}, 
\notag\\
& \hphantom{\sum_{(u,v)\in E}\phi_{(u,v)} = \ } \forall v \in V \setminus \{s,t\}, 
\label{eq:spp_flow}\\
& (x_u,x_v) \in \mathcal{X}_e, \quad \forall e=(u,v)\in\mathcal{E}_p .
\label{eq:spp_edge_constraint}
\end{align}
\end{subequations}

Here, the objective is to minimize the cost of traversing the path $p$, with $c(x_u, x_v)$ denoting the cost of each edge $e := (u, v)$. The binary variables $\phi_e$ select which edges belong to the optimal path $p$, and the chosen edges form the set $\mathcal{E}_p$. Constraints~\eqref{eq:spp_endpoints} and~\eqref{eq:spp_containment} confine each vertex to its convex set, with the endpoint configurations placed in the source set $\mathcal{C}_s$ and target set $\mathcal{C}_t$. As in typical SPP formulations~\cite{nonlinear_programming}, constraints~\eqref{eq:spp_injection} and~\eqref{eq:spp_flow} route one unit of flow from the source $s$ to the target $t$ while conserving flow at every intermediate vertex. Finally, constraint~\eqref{eq:spp_edge_constraint} enforces that the trajectory $x$ satisfies all edge continuity constraints along the path $p$. 

As presented, Problem~\eqref{eq:spp_in_gcs} is a mixed-integer program that can be solved to global optimality via branch and bound techniques~\cite{gcs_optimiazation_paper_original}. However, this approach does not scale well to several motion planning problems of practical interest~\cite{gcs_speedup_gcs_star}. 

A key result from~\cite{gcs_planning_paper_original} is that, instead of solving Problem~\eqref{eq:spp_in_gcs} directly, we can solve a tractable convex relaxation of the underlying mixed-integer program. This relaxation replaces each binary $\phi_e$ with a continuous value $\phi_e \in [0, 1]$. Rather than isolating a single discrete path, these fractional values generate a probability distribution over all possible paths. This acts as a global guide, weighting edges based on their likelihood of belonging to the optimal path. We can therefore sample candidate paths from $s$ to $t$ according to these probabilities. With $\phi_e$ now fixed for a given sample, a rounding procedure recovers a feasible, near-optimal solution. Specifically, Problem~\eqref{eq:spp_in_gcs} reduces to a convex program over the continuous variable $x$, which is solved for each sampled path before returning the one with the lowest cost.

While far more efficient than solving Problem~\eqref{eq:spp_in_gcs} exactly, the convex relaxation is still costly for large graphs~\cite{gcs_speedup_gcs_star}. Additionally, the cost of evaluating the sampled paths grows linearly with the number of samples. The following sections address these two computational bottlenecks. 

\section{Data-Driven Acceleration of GCS} \label{sec:data_driven_acceleration_of_gcs}
The above procedure yields solutions with a certifiably small optimality gap for Problem~\eqref{eq:spp_in_gcs}~\cite{gcs_planning_paper_original}. We leverage this property to generate an offline library of high-quality GCS solutions. We generate the dataset by randomizing both the planning environment, which defines the convex sets $\mathcal{C}$, and the start and target configurations that make up the global query $\mathcal{Q} = (q_{\text{initial}}, q_{\text{target}})$. Solving the corresponding GCS instance produces the relaxation flow vector $\Phi \in [0, 1]^{|E|}$, where $|\cdot|$ denotes set cardinality, alongside a set of candidate paths $\mathcal{P}$ ranked by optimality. Accumulated across $K$ randomized instances, these solutions serve as our training targets. An overview of this process is shown in Fig.~\ref{fig:training_diagram}.  In the following subsections, we detail how this dataset is used to replace most of the computationally expensive online optimization steps with learned neural network surrogates.
\vspace{-0.5mm}
\subsection{Learned Convex Relaxation} 
\label{sec:learned_convex_relaxation}
As we shall see in Section \ref{sec:experiments_and_results}, the convex relaxation optimization problem can take several seconds to solve for various motion planning tasks. As a result, it struggles to meet the speed requirements of online replanning. To accelerate this process, we rely on the fact that the solution to any given GCS instance depends on the local structure of the graph, represented by the convex sets $\mathcal{C}_v$, and the global query $\mathcal{Q}$. Our objective in this subsection, is to train a neural network conditioned on these exact parameters to predict the convex relaxation output. This effectively replaces the costly online optimization step with a rapid neural network forward pass.

As the first step, a feature multilayer perceptron (MLP) processes the convex set representations, such as the obstacle-free halfspaces used in collision-free motion planning. Attention pooling aggregates these features into a vector, $\mathcal{F}$. To ensure each node carries the global context of the planning task, a conditioning MLP fuses $\mathcal{F}$ with the query $\mathcal{Q}$ to initialize the node embeddings. These query-seeded embeddings are subsequently passed through Bidirectional Graph Attention (BiGATv2) layers \cite{graph_attention}. Within these layers, dynamic attention weights are computed directly from the embeddings, allowing the network to prioritize adjacent nodes that form viable directed pathways to the target. To account for the directed nature of the GCS graph, two independent attention streams process incoming and outgoing edges, enabling each node to aggregate context from both its predecessors and successors. Stacking four such layers allows each node to progressively incorporate information from surrounding convex sets up to four hops away, yielding the context-aware node embeddings, $\mathcal{N}$. Finally, an edge logit MLP processes the embeddings $\mathcal{N}$ and the query $\mathcal{Q}$. A sigmoid activation function maps this output to the predicted edge flows stacked into the vector $\hat{\Phi}_{\text{predicted}}$. These predictions are supervised via a binary cross entropy with logits loss against the ground-truth vector $\Phi$. 

These $\hat{\Phi}_{\text{predicted}}$ do not inherently satisfy the global flow conservation constraints required for a valid path. To resolve this, we project the network's output onto the feasible flow space using a lightweight convex quadratic program. The objective function minimizes the deviation from the learned predictions. The constraints utilize a node-edge incidence matrix to enforce strict flow conservation. This requires the net flow to be zero at all intermediate vertices, while injecting exactly one unit of flow at the source node and extracting it at the target node. Thus, the mathematical program is:
\begin{equation}
\label{eq:projection}
\begin{aligned}
\hat{\Phi} &= \operatorname*{arg\,min}_{\Phi' \in [0,1]^{|E|}} \tfrac{1}{2}\left\|\Phi' - \hat{\Phi}_{\text{predicted}}\right\|_2^2 \\
&\phantom{{}={}} \text{s.t.}\ \ B\Phi' = b,\ \ B \in \mathbb{R}^{|V| \times |E|},\ \ b \in \mathbb{R}^{|V|}, \\
B_{v,e} &= \begin{cases}
  +1 & \text{if } e \text{ leaves } v \\
  -1 & \text{if } e \text{ enters } v \\
  \phantom{+}0 & \text{otherwise}
\end{cases}
\quad
b_v = \begin{cases}
  +1 & \text{if } v = s \\
  -1 & \text{if } v = t \\
  \phantom{+}0 & \text{otherwise}
\end{cases}
\end{aligned}
\end{equation}
\subsection{Ranking-Driven Solution Rounding}
\label{sec:ranking_driven_rounding}
While the learned relaxation stage efficiently estimates the projected flows $\hat{\Phi}$, recovering a feasible solution still requires solving the underlying trajectory optimization for the continuous variables $x$ during the rounding stage. Nominal GCS typically evaluates all candidate paths sampled from the flow distribution, which is computationally demanding. To circumvent this, we introduce a learned ranking mechanism that orders these candidate paths based on the predicted flows. If this ranking is accurate, we can simply evaluate the top-ranked paths, verify their feasibility, and terminate the search early. This early-exit capability provides significant computational speedups. In the worst-case scenario, where the ranking is less accurate, the algorithm systematically evaluates the remaining paths, reverting to the standard computation time of nominal GCS with only a negligible neural network inference overhead.

We train the ranker as follows. For a candidate path $p_m := (s, v_0, \dots, v_k, \dots, t)$ drawn from the GCS training data, we gather the context-aware node embeddings $\mathcal{N}$ and projected flows $\hat{\Phi}$. At each step $k$, we form a token by concatenating the node embedding $h(v_k) \in \mathcal{N}$, the incoming edge flow $\hat{\phi}_{(v_{k-1}, v_k)} \in \hat{\Phi}$, and an 8-dimensional sinusoidal positional encoding standard in transformer architectures~\cite{attention}. A Transformer encoder processes this token sequence, using self-attention over the trajectory and padding masks for variable-length paths, capturing routing patterns along the path. Masked mean pooling then condenses the sequence into a path-level vector, which a score MLP maps to a scalar score $s(p_m)$.

We supervise these scores with a pairwise ranking loss. From the candidate paths $\mathcal{P}$ ordered by cost, we form the set $\mathcal{R}$ of pairs $(p_i, p_j)$ in which $p_i$ incurs a lower total trajectory cost than $p_j$ based on Problem~\ref{eq:spp_in_gcs}. We define this  cost gap as:$$\Delta c_{ij} = \sum_{e:=(y,z) \in \mathcal{E}_{p_j}} c(x_y, x_z) - \sum_{e:=(u,v) \in \mathcal{E}_{p_i}} c(x_u, x_v) > 0$$The network is trained to score the cheaper path higher by minimizing the following RankNet loss:$$\mathcal{L} = \frac{1}{\vert{}\mathcal{R}\vert{}}\sum_{(i,j)\in\mathcal{R}} \left( -\bar{P}_{ij} o_{ij} + \log(1 + e^{o_{ij}}) \right)$$Here, $\vert{}\mathcal{R}\vert{}$ is the number of path ordered pairs, and the predicted score margin is $o_{ij} = s(p_i) - s(p_j)$. The term $-\bar{P}_{ij} o_{ij}$ drives the network to maximize the predicted margin $o_{ij}$. The soft target $\bar{P}_{ij} = \sigma(\Delta c_{ij}/\tau)$ scales this push based on the true cost gap, where $\sigma$ is the sigmoid function and $\tau$ is the temperature parameter. Small gaps where $\bar{P}_{ij} \approx 0.5$ gently score near-equivalent paths similarly, while large gaps where $\bar{P}_{ij} \to 1$ strictly separate optimal paths from costly alternatives. The regularizer term $\log(1 + e^{o_{ij}})$ prevents the network from making these scores arbitrarily large. At runtime, the learned scores $s(p_m)$ order the candidate paths so the planner evaluates the most promising paths first.
\begin{figure*}[t]
\centering
\includegraphics[width=0.95\textwidth]
{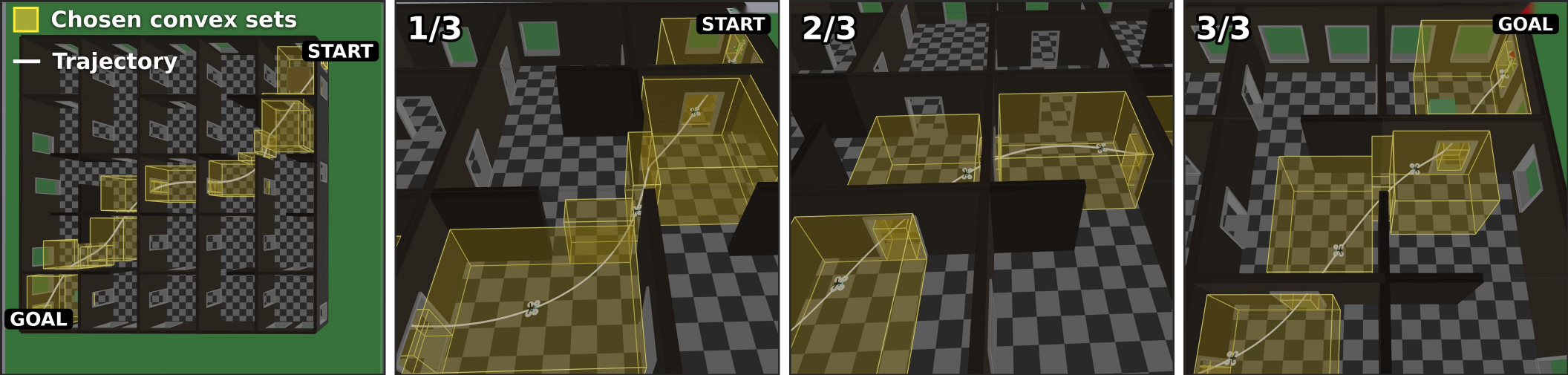}
\caption{A collision-free trajectory for a 3D quadrotor navigating a building interior. The leftmost panel provides a top-down overview of the complete path from start to goal, while the remaining panels show sequential close-up views of the flight.}
\label{fig:quadrotor}
\end{figure*}
\begin{table*}[t]
\caption{Mean computation time and path cost comparisons for 3D quadrotor motion planning across 100 test instances.}
\label{tab:quadrotor}
\centering
\setlength{\tabcolsep}{5pt}
\footnotesize
\begin{tabular}{@{}l ccccc ccccc@{}}
\toprule
& \multicolumn{5}{c}{Convex} & \multicolumn{5}{c}{Non-Convex} \\
\cmidrule(lr){2-6} \cmidrule(l){7-11}
Method & Paths & CR/GAT (s) & Rounding (s) & Total (s) & $C_{\text{round}}$
       & Paths & CR/GAT (s) & Rounding (s) & Total (s) & $C_{\text{round}}$ \\
\midrule
Nominal GCS            & 10.0 & 28.564 (CR) & 0.226 & 28.790 & \textbf{13.780} & 10.0 & 6.821 (CR)  & 1.406 & 8.227 & \textbf{18.128} \\
Neural GCS w/o RankNet & 10.0 & \textbf{0.005} (GAT) & 0.227 & 0.232  & 13.811 & 10.0 & \textbf{0.007} (GAT) & 1.416 & 1.423 & 18.143 \\
Neural GCS w/ RankNet  & \textbf{1.0} & \textbf{0.005} (GAT) & \textbf{0.022} & \textbf{0.027}  & 17.386 & \textbf{1.0} & \textbf{0.007} (GAT) & \textbf{0.157} & \textbf{0.164} & 26.113 \\
\midrule
FastPathPlanning~\cite{quadrotor_fast_planning_tobia}       & --  & --          & --    & 0.069  & --     & --  & --          & --    & --    & --     \\
\bottomrule
\end{tabular}
\end{table*}
\section{Experiments and Results}
\label{sec:experiments_and_results}
We refer to our learning-based pipeline for accelerating motion planning using GCS as Neural GCS. This section presents empirical evaluations of Neural GCS across diverse robotic systems. We further investigate the tradeoff between computation time and solution optimality by benchmarking Neural GCS against both the nominal GCS formulation and recent domain-specific GCS acceleration strategies.
\subsection{Experimental Setup}
\label{sec:experimental_setup}
\subsubsection{Data Collection and Model Training}
\label{sec:dataset_collection_and_model_training}
For all experiments in this section, we use nominal GCS to generate 500 training, 100 validation, and 100 test instances. We rely on implementations from~\cite{gcs_planning_paper_original,gcs_nonconvex_paper} for collision-free motion planning and~\cite{gcs_planning_through_contact} for planning through contact, making no additional attempts to improve the runtime performance of these baselines. Training is conducted offline on a single NVIDIA GeForce RTX 3070 Ti Laptop GPU. Following the two phases in Fig.~\ref{fig:training_diagram}, we train the flow prediction network and freeze its weights so that the node embeddings it produces serve as inputs to the path ranking network. Generating the offline datasets and training the models for all problem variants presented in this section takes approximately 25 hours combined, which is a modest computational requirement.

\subsubsection{Benchmarks and Evaluation Metrics}
\label{sec:benchmarks_and_evaluation_metrics}
To isolate the contribution of each component, we consider two variants of our method. The first, Neural GCS without path ranking, evaluates all candidate paths produced by the flow prediction network. The second utilizes our complete method, running both the flow prediction network and the ranking-driven rounding mechanism. The original GCS formulation of Marcucci et al.~\cite{gcs_planning_paper_original} serves as our primary baseline. Generating smoother trajectories, however, introduces non-convex constraints like true acceleration limits. To fit these problems into the GCS paradigm, \cite{gcs_nonconvex_paper} utilizes convex surrogates for these constraints during the convex relaxation stage. Once the path is fixed, the rounding stage solves the exact non-convex programs to recover a feasible trajectory. We benchmark our approach against such non-convex extensions of~\cite{gcs_planning_paper_original} as well.

We evaluate all metrics on the 100 test instances and measure planning time on an Intel Core i9-14900K CPU. Inference requires no GPU, making the pipeline suitable for robots with limited onboard compute. Across all experiments, we report the computation times broken down into two distinct stages. For nominal GCS, the convex relaxation (CR) timings correspond to the time required to solve the convex optimization problem. For Neural GCS, the reported GAT timings cover both the forward pass through the learned convex relaxation module and the flow conservation QP that follows it. Finally, the rounding timings for both methods encompass the time spent sampling candidate paths and solving the exact trajectory optimization for the evaluated candidates. Every method achieves a 100\% success rate across all tasks, so we omit this metric from further reporting.
\begin{figure*}[t]
\centering
\includegraphics[width=0.9\textwidth]
{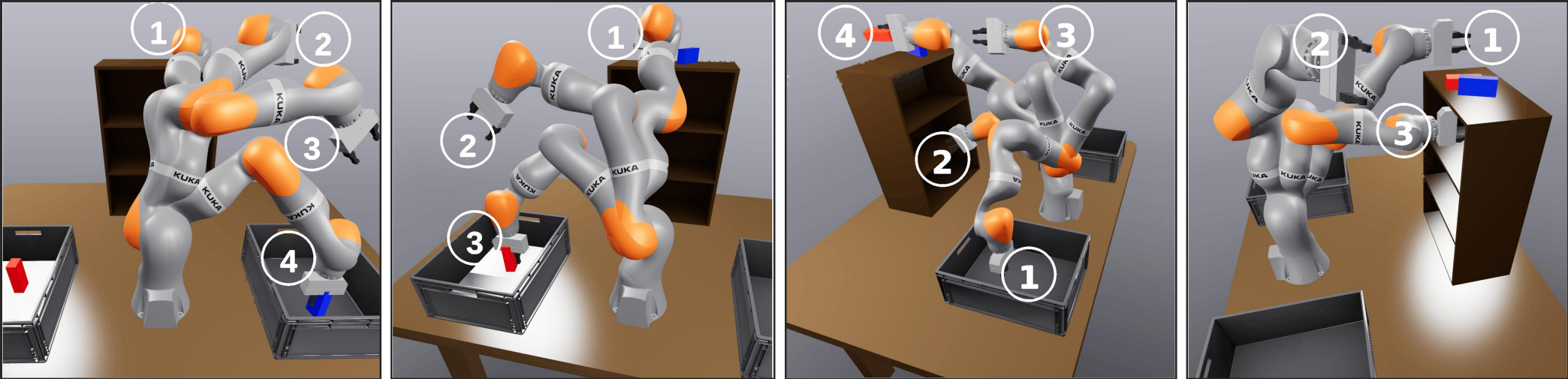}
\caption{A multi-step pick-and-place sequence for a 7-DoF manipulator. From left to right: (1) moving to pick up the blue block, (2) placing the blue block and reaching for the red block, (3) placing the red block, and (4) returning to the middle of the shelf.}
\label{fig:manipulation}
\end{figure*}
\begin{table*}[t]
\caption{Mean computation time and path cost comparisons for 7-DoF arm motion planning across 100 test instances.}
\label{tab:manipulator}
\centering
\setlength{\tabcolsep}{5pt}
\footnotesize
\begin{tabular}{@{}l ccccc ccccc@{}}
\toprule
& \multicolumn{5}{c}{Convex} & \multicolumn{5}{c}{Non-Convex} \\
\cmidrule(lr){2-6} \cmidrule(l){7-11}
Method & Paths & CR/GAT (s) & Rounding (s) & Total (s) & $C_{\text{round}}$
       & Paths & CR/GAT (s) & Rounding (s) & Total (s) & $C_{\text{round}}$ \\
\midrule
Nominal GCS            & 10.0 & 0.192 (CR) & 0.114 & 0.306 & \textbf{4.190} & 10.0 & 5.637 (CR) & 1.996 & 7.633 & \textbf{6.771} \\
Neural GCS w/o RankNet & 10.0 & \textbf{0.003} (GAT) & 0.113 & 0.116 & 4.197 & 10.0 & \textbf{0.003} (GAT) & 1.876 & 1.879 & 6.849 \\
Neural GCS w/ RankNet  & \textbf{1.0} & \textbf{0.003} (GAT) & \textbf{0.007} & \textbf{0.010} & 4.235 & \textbf{1.0} & \textbf{0.003} (GAT) & \textbf{0.130} & \textbf{0.133} & 7.456 \\
\midrule
Multi-Query GCS~\cite{gcs_speedup_savva_offline_sdp}        & -- & -- & -- & 0.016 & 4.441 & -- & -- & -- & -- & -- \\
\bottomrule
\end{tabular}
\end{table*}
\vspace{-5mm}
\subsection{Collision-Free Navigation for a 3D Quadrotor}
\label{sec:collision_free_navigation_3d_quadrotor}
Fig.~\ref{fig:quadrotor} demonstrates our approach for a 3D quadrotor navigating a cluttered building interior, with rooms connected via narrow doors and windows. Exploiting differential flatness~\cite{differential_flatness}, we reduce the planning problem to optimizing a 3D position trajectory. The obstacle-free regions are modeled as axis-aligned bounding boxes, which serve as the sets $\mathcal{C}_v$. To synthesize the training dataset, we use a $20\text{ m} \times 20\text{ m}$ environment, randomizing the locations of walls, doors, and windows along with the 3D start and goal poses. We then test the pipeline on $15\text{ m} \times 15\text{ m}$ and $25\text{ m} \times 25\text{ m}$ building layouts. These layouts yield graphs both smaller and larger than those seen during training, demonstrating that the network generalizes across graph sizes.

We parameterize the continuous trajectory $x$ in Problem 1 as a piecewise Bézier curve, constrained to be $C^4$-continuous across the traversed convex regions. Our benchmarks in Table~\ref{tab:quadrotor} include the original convex GCS formulation\cite{gcs_planning_paper_original}, its non-convex variant enforcing true acceleration limits~\cite{gcs_nonconvex_paper}, and FastPathPlanning\cite{quadrotor_fast_planning_tobia}, a heuristic planner tailored for rapid path computation through safe axis-aligned boxes.

As shown in Table~\ref{tab:quadrotor} nominal convex GCS spends 28.564 seconds on the relaxation and nominal non-convex GCS spends 6.821 seconds. This difference in CR solve times stems from the underlying B\'{e}zier parameterizations. Specifically, the non-convex variant forms a smaller SOCP by using order-6 control points instead of the order-7 convex formulation. Neural GCS recovers similar edge flows much faster, in 0.005 and 0.007 seconds, respectively.

Without ranking-driven rounding, Neural GCS evaluates the same average number of candidates as nominal GCS, and its rounding times are correspondingly near-identical. With ranking-driven rounding, the planner reaches a feasible path after a single trajectory optimization. Rounding time falls from 0.227 to 0.022 seconds in the convex setting and from 1.416 to 0.157 seconds in the non-convex setting, yielding an order of magnitude reduction in both cases.

Finally, when comparing against FastPathPlanning, we omit its stage-wise timings and trajectory cost because it follows an entirely different algorithm, and restrict the comparison to total computation time. FastPathPlanning completes in 0.069 seconds on average, while Neural GCS with ranking-driven rounding requires 0.031 seconds in the convex setting and 0.168 seconds in the non-convex setting. We thus match FastPathPlanning in computation speed, but our method can additionally tackle planning in configuration space and the contact-rich problems we turn to next.
\begin{figure*}[t]
\centering
\includegraphics[width=0.9\textwidth]
{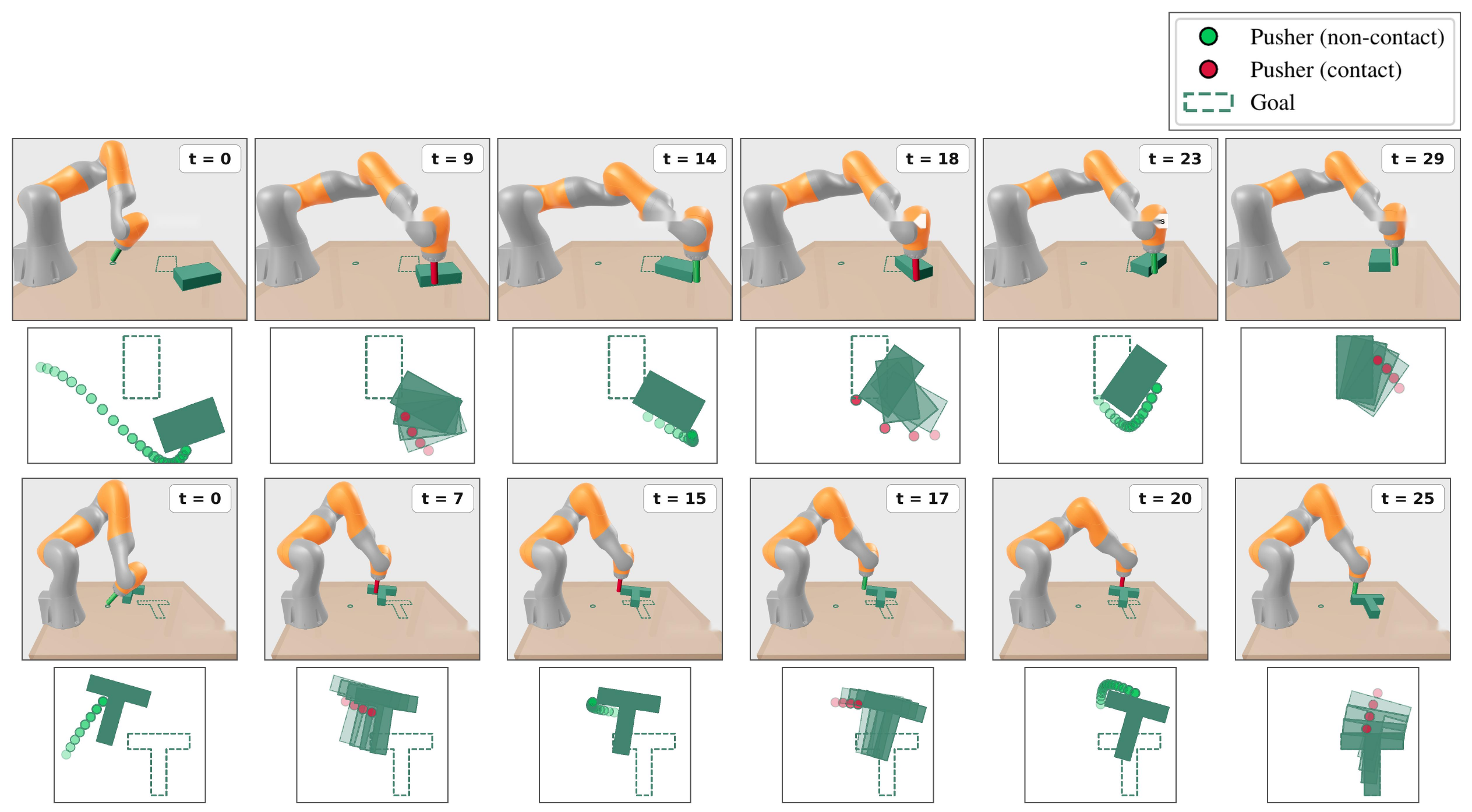}
\caption{Planar pushing sequences for a rectangular box shaped object (top two) and a T-shaped object (bottom two). Each sequence displays the 3D manipulator poses over time alongside corresponding top-down 2D views of the object and pusher trajectories.}
\label{fig:planning_through_contact}
\end{figure*}
\begin{table*}[t]
\caption{Mean computation time and path cost comparisons for planning through contact across 100 test instances.}
\label{tab:pushing}
\centering
\setlength{\tabcolsep}{5pt}
\footnotesize
\begin{tabular}{@{}l ccccc ccccc@{}}
\toprule
& \multicolumn{5}{c}{Box} & \multicolumn{5}{c}{Tee} \\
\cmidrule(lr){2-6} \cmidrule(l){7-11}
Method & Paths & CR/GAT (s) & Rounding (s) & Total (s) & $C_{\text{round}}$
       & Paths & CR/GAT (s) & Rounding (s) & Total (s) & $C_{\text{round}}$ \\
\midrule
Nominal GCS            & 100.0 & 7.476 (CR) & 9.773 & 17.249 & \textbf{74.33} & 100.0 & 88.062 (CR) & 30.183 & 118.245 & \textbf{68.01} \\
Neural GCS w/o RankNet & 100.0 & 0.004 (GAT) & 3.678 & 3.682 & 75.03 & 100.0 & 0.006 (GAT) & 7.158 & 7.162 & 71.18 \\
Neural GCS w/ RankNet  & \textbf{5.8} & \textbf{0.003} (GAT) & \textbf{0.306} & \textbf{0.309} & 90.76 & \textbf{8.0} & \textbf{0.004} (GAT) & \textbf{0.607} & \textbf{0.611} & 90.62 \\
\bottomrule
\end{tabular}
\vspace{-4mm}
\end{table*}
\subsection{Multi-Step Pick-and-Place with a 7-DoF Manipulator}
Fig.~\ref{fig:manipulation} demonstrates our approach on a 7-DoF manipulator performing a multi-step pick-and-place task. This involves relocating a red and a blue block from two adjacent bins to the top of a shelf. We formulate this entire process as a sequence of motions navigating between pre-pick, pick, pre-place, and place poses, all of which are planned using GCS. We plan directly in the 7-DoF joint configuration space, ensuring the recovered trajectory respects the arm's kinematics and joint limits without a separate inverse-kinematics stage. The collision-free configuration space is modeled as a union of convex polytopes, which serve as the sets $\mathcal{C}_v$. To synthesize the training dataset, we randomize the placement of the shelf and bins along with the start and goal configurations.

Table~\ref{tab:manipulator} highlights our computational advantage for both the original convex GCS formulation~\cite{gcs_planning_paper_original} and its non-convex extension~\cite{gcs_nonconvex_paper}. Nominal convex and non-convex GCS spend 0.192 and 5.637 seconds on the relaxation, respectively. This timing gap stems from their parameterizations: the convex setting uses a single control point per region, while the non-convex variant enforces joint limits via higher-order B\'{e}zier curves, yielding a substantially larger SOCP. Bypassing this SOCP entirely, Neural GCS recovers similar edge flows in just 0.003 seconds for both settings. With ranking-driven rounding, the planner reaches a feasible path after an average of just one trajectory optimization. Consequently, rounding time drops from 0.114 to 0.007 seconds in the convex setting, and from 1.996 to 0.130 seconds in the non-convex setting.

We further benchmark against multi-query GCS~\cite{gcs_speedup_savva_offline_sdp}, which solves one semidefinite program (SDP) per goal region to yield a certified lower bound on the cost-to-go. Online, it sequentially solves small convex programs to score the best neighboring set until reaching the goal. Once this sequence of sets is fixed, a final convex program recovers the trajectory for the resulting path. Because~\cite{gcs_speedup_savva_offline_sdp} derives these bounds only for single-control-point parameterizations, our comparison applies solely to the convex variant. Across the 100 test instances, online planning averages 0.016 seconds at a cost comparable to Neural GCS. However, the primary drawback is the offline SDP stage, which takes two to three minutes and must be recomputed for every new GCS layout, adding substantial computational overhead.

\subsection{Planning Through Contact for Planar Pushing}
Finally, we consider planning through contact, where a mechanical pusher must maneuver a sliding object to its target pose. Here, simultaneously optimizing the contact wrench and locations introduces non-convex equality constraints. To address this, the nominal GCS formulation proposed by Graesdal et al.~\cite{gcs_planning_through_contact} utilizes tight semidefinite relaxations where the convex sets paired with contact-mode vertices are modeled as spectrahedra. The non-contact modes represent the pusher moving through collision-free space to transition between contact faces. Because these modes are governed by linear constraints, their paired graph vertices are represented as polyhedra. We note that while domain-specific GCS accelerators exist, such as FastPathPlanning for quadrotors and Multi-Query GCS for manipulation, no such GCS techniques are currently available for planning through contact. Therefore, we utilize the nominal planner both as a baseline and as the expert data generation mechanism to train our learning-based method. We evaluate our approach on two distinct slider geometries: a rectangular box with four faces and a T-shaped object with eight faces. Fig.~\ref{fig:planning_through_contact} illustrates the successful generation of long-horizon motion plans.

To clearly distinguish between the contact and non-contact modes within our learning-based abstraction, we describe every GCS node using a unified feature vector that intuitively summarizes its underlying geometry. For contact modes, this vector comprises the geometric parameters of the active slider face, specifically its midpoint and length. For non-contact modes, it encodes the polygonal boundary of the corresponding collision-free space. This representation comprises the node feature vector $\mathcal{F}$, which is used to calculate the node embeddings $\mathcal{N}$ (Fig.~\ref{fig:training_diagram}, middle).

Table~\ref{tab:pushing} details the computation times and path costs for planar pushing. For Nominal GCS, the CR stage creates a significant bottleneck, requiring 7.476 seconds for the Box and 88.062 seconds for the Tee. Neural GCS generates these edge flows in just 0.004 and 0.006 seconds, respectively. With ranking-driven rounding, the planner finds a feasible solution after evaluating an average of only 5.8 paths for the Box and 8.0 for the Tee. Ultimately, the full Neural GCS pipeline reduces total computation time from 17.249 to 0.309 seconds for the Box, and from 118.245 to 0.611 seconds for the Tee, achieving a two-orders-of-magnitude speedup.

\subsection{Discussion and Limitations}
One of the primary challenges with nominal GCS is the rapidly expanding size of the CR optimization problem for larger graphs. For example, the SDP for the Tee pushing problem involves roughly 200,000 constraints, resulting in solve times of approximately 90 seconds~\cite{gcs_planning_through_contact}. In fact, for objects exceeding eight faces, we were unable to fit the resulting SDP into CPU memory. This scaling issue can be mitigated to some extent by leveraging the sparsity structure inherent in optimal control problems~\cite{sparsity_in_sdps}. In contrast, while larger graphs naturally require our  approach to process more node features, this feedforward operation scales significantly better than solving massive mathematical programs. As a result, our method executes the learned CR stage in under 10 milliseconds across all problem instances.

As for the rounding stage, the path ranking network successfully reduces the number of candidate paths evaluated online. While this early termination strategy does achieve a 100\% success rate, it does not always prioritize the least-cost path. This is evidenced by the higher $C_{\text{round}}$ values across all evaluated robotic domains. This result highlights a fundamental tradeoff of our approach where some loss in optimality yields a substantial computational speedup. While increasing the dataset size and expanding the network architecture could potentially mitigate this gap, it would come at the expense of slightly increased inference times. Additionally, our approach introduces a structural limitation for contact-rich tasks. If the underlying SDP becomes prohibitively expensive to solve offline, we cannot generate the corresponding ground-truth dataset required to train Neural GCS.

Finally, a useful extension of our method involves planning under uncertainty via chance constraints. For our quadrotor and manipulator examples, obstacle boundaries can be directly inflated as a closed-form function of sensing uncertainty~\cite{motion_planning_2}. This computationally inexpensive step modifies the convex sets prior to planning, ensuring a larger collision-free buffer when the robot is less certain of its surroundings. Because our network successfully generalizes to unseen convex sets, this robust formulation would directly benefit from our approach's fast computation times.

\section{Conclusion}
In this work, we introduced Neural GCS, a data-driven pipeline designed to accelerate mixed discrete-continuous motion planning. We leveraged nominal GCS as an offline ground-truth data generator and replaced its computationally expensive online optimization steps with lightweight neural network surrogates. We demonstrated the efficacy of this approach across a wide variety of robotic systems without requiring any domain-specific tuning. For the broad class of problems that nominal GCS can solve, our framework provides significant computational acceleration, generating solutions fast enough to support online replanning rates, thus establishing the generality of Neural GCS.

\bibliographystyle{IEEEtran}
\bibliography{IEEEabrv,references}

\end{document}